\pdfoutput=1
\documentclass[11pt]{article}

\usepackage[preprint]{acl}

\usepackage{times}
\usepackage{latexsym}
\usepackage{listings}
\usepackage{multirow}
\usepackage{multicol}
\usepackage{tabularx}
\usepackage{authblk}

\usepackage[T1]{fontenc}
\usepackage[utf8]{inputenc}

\usepackage{microtype}

\usepackage{inconsolata}

\usepackage{graphicx}

\title{Simulating User Diversity in Task-Oriented Dialogue Systems \\using Large Language Models}

\author{Adnan Ahmad \and Stefan Hillmann \and Sebastian Möller\\ \\
Quality and Usability Lab, Technische Universität Berlin \\
\href{mailto:adnan.ahmad@tu-berlin.de}{adnan.ahmad@tu-berlin.de}, \href{mailto:stefan.hillmann@tu-berlin.de}{stefan.hillmann@tu-berlin.de}, \href{mailto:sebastian.moeller@tu-berlin.de}{sebastian.moeller@tu-berlin.de}}

\begin{document}
\maketitle
\begin{abstract}
In this study, we explore the application of Large Language Models (LLMs) for generating synthetic users and simulating user conversations with a task-oriented dialogue system and present detailed results and their analysis. We propose a comprehensive novel approach to user simulation technique that uses LLMs to create diverse user profiles, set goals, engage in multi-turn dialogues, and evaluate the conversation success. We employ two proprietary LLMs, namely GPT-4o and GPT-o1 \cite{openai2024gpt4}, to generate a heterogeneous base of user profiles, characterized by varied demographics, multiple user goals, different conversational styles, initial knowledge levels, interests, and conversational objectives. We perform a detailed analysis of the user profiles generated by LLMs to assess the diversity, consistency, and potential biases inherent in these LLM-generated user simulations. We find that GPT-o1 generates more heterogeneous user distribution across most user attributes, while GPT-4o generates more skewed user attributes. The generated set of user profiles are then utilized to simulate dialogue sessions by interacting with a task-oriented dialogue system.

\end{abstract}

\section{Introduction}

Task-oriented dialogue systems are specialized to achieve custom defined goals by retrieving information and generating responses from internal sources of data. Contrary to today's general purpose Large Language Model (LLM)-based chatbots, which are more robust in terms of conversation capabilities (\citet{brown2020language}, \citet{wei2021finetuned}) but often lack domain knowledge, internal access to data, and resource constraints, task-oriented dialogue systems are often constrained by the amount of initial user data required to train, evaluate, as well as to further design and improve the system. 

We propose a user simulation design for task-oriented dialogue systems using LLMs and generate experimental data by interacting the simulated users with a task-oriented dialogue system. Our user simulation approach is greatly influenced by the  ideas presented in Schatzmann and Young (2009) \cite{schatzmann2009hidden}, which emphasizes the importance of a structured, goal-oriented user simulation. Our approach is to create detailed and diverse user profiles and user behavior specifications by extending their proposed template to represent user goals. By integrating structured elements into our synthetic user profile generation, we ensure that the dialogue system can be effectively simulated by managing and responding to complex user interactions that are required to interact with a task-oriented dialogue system. 

We use GPT-4o\footnote{\url{https://platform.openai.com/docs/models\#gpt-4o}} and GPT-o1\footnote{\url{https://platform.openai.com/docs/models\#o1}} to generate the synthetic user profile with a diverse set of attributes. Afterwards, we simulate the conversation using the generated users with a task-oriented chatbot. This approach allows us to largely eliminate human involvement in the simulation pipeline, user profile creation, conversation simulation, and evaluation of task success, thereby achieving a more automated system. Additionally, we conduct a comparative analysis of the user profiles generated by GPT-4o and GPT-o1 to examine their comparative characteristics in the context of the generated user attributes. GPT-4o is a chat model, while GPT-o1 is a reasoning model. We show  how the generated users widely vary based on the model capabilities, despite having an equal prompt and user template. This analysis helps us to understand the generation of realistic and diverse user profiles and the related simulated interactions.

Incorporating diversity in the properties of simulated users is pivotal for the robustness and comprehensiveness of user simulation. By creating a heterogeneous user base with varied demographics, objectives, and conversational styles, we aim to mirror the complex and multifaceted nature of real-world interactions. This diversity not only tests the system’s ability to handle a wide range of inquiries and interactions effectively, but also ensures that the dialogue system is resilient across different user types, user goals, interests, and scenarios. By analyzing how the differences in the simulated user behavior influence the system's performance, we can identify potential areas for improvement, ensuring the system is equitable and effective for all user groups. 

For the dialogue simulation purpose, we develop StudyBot, a task-oriented chatbot integrated with a comprehensive real-time database containing multiple attributes related to the wide range of study programs offered by Technische Universität Berlin (TUB). The primary objective of the StudyBot chatbot is to assist prospective students by delivering pertinent information regarding the study programs available at the university. This is achieved through a multi-turn dialogue system that processes and responds to users’ inquiries and interests in natural language. The StudyBot has all the components of a modular task-oriented dialogue system \cite{zhao2019review}. In addition, we use an open-weights LLM, namely Mistral-7B-Instruct-v0.2 \cite{jiang2023mistral}, for in-context natural language response generation.  We perform 57 such dialogue sessions with diverse generated users, each containing \textasciitilde7 user turns on average, and StudyBot achieves 82.46\% task success in terms of successfully attaining defined user goals for each of these conversations. We report the details of the design of such a user simulator and present the results of the user simulation based on various user attributes, diversity, bias, and goals. Such a pipeline enables us to generate more data for further use to evaluate and improve the current chatbot, which is often a bottleneck for many task-oriented dialogue systems. 

\section{Related Studies}

User simulation is being used more often by task-oriented dialogue systems to test and develop dialogue techniques (\citet{gur2018user}). The Hidden Agenda Model, a user simulation that uses an organized method to replicate real-world user behaviors, was first presented by \citet{schatzmann2009hidden}. Our user simulation approach is greatly influenced by the approach presented in \citet{schatzmann2009hidden}. This is an influential work on modern dialogue systems, as well as user simulation. This model emphasizes the importance of a structured, goal-oriented user simulation, which has shaped our approach of creating detailed user profiles and behavior specifications utilizing a template. By integrating these structured elements into our user simulation, we ensure that the dialogue system can effectively manage and respond to complex user interactions in a task-oriented manner. Moreover, LLMs are integrated with these user simulations to improve the dialog system's capacity to produce responses that are human-like and to dynamically adjust to user preferences (\citet{abdullin2024synthetic}, \citet{hudevcek2023large}).

\section{Dataset}

Our task-oriented dialogue system is designed based on the study program database of TUB\footnote{\url{https://www.tu.berlin/en/studying/study-programs/all-programs-offered}}. We were provided API access to the database which contains all the information related to all study programs offered at TUB. There are in total 147 study programs, with a variety of degree types, admission requirements, etc. There are 16 different attributes in the database (e.g., program description, ECTS points, admission requirements, etc.). We create a relational database using the attributes which the chatbot will use to retrive information from. We define the intents and entity classes from these attributes. The NLU part of our task-oriented dialogue system uses these attributes and their corresponding values to train an intent and entity recognition model. We use the Rasa's default DIETClassifier\footnote{\url{https://rasa.com/docs/rasa/reference/rasa/nlu/classifiers/diet_classifier/}} for NLU training. We also incorporate semantic search to enhance results based on multilingual embeddings on top of the classifier.

\section{Experiment Setup}

\subsection{User Template Design}

For the user simulation experiment setup, we first create a user definition template in a JSON format. The template contains the metadata, user demographic info, behavioral attributes, interests, and goals. The template has fixed values as well as dynamic values. The fixed values are, for example, primary goal (e.g., to find relevant study programs), user role (e.g., prospective student) etc. The rest of the values are set by the LLMs for each generated user profile. Some values are set by the LLM in a free choice manner (e.g., general interests), others are set by the LLM from predefined options (e.g., gender, language preference, goals, etc.). An example of a generated user with complete JSON structure is given in figure \ref{fig:user} in Appendix \ref{sec:appendix}. We use the exact same prompt template to generate 100 user profiles using both GPT-4o and GPT-o1, so that the generated users are directly comparable. 

The prompt is dynamic and it consists of a system prompt, statistics of already generated users (attributes and their value counts), and a user template to fill in. The system prompt and the template remain the same across different LLMs. We instruct to generate diverse users, but to reflect a real-world distribution in the generation. A detailed analysis between the generated users of these two models is given in the result and discussion section.

\subsection{Chat-bot Interaction}

Once the users are created, we use them to interact with the StudyBot by initializing a conversation. For that, there is an initial prompt given to the LLM with the implemented user template throughout the conversation (as system prompt). The previous conversation history is also added for each response generation. For selecting options through buttons given by StudyBot, we use another special prompt that ensures the format of the button's selection text. The user utterances are generated using GPT-4o model (given the user template and a system prompt). The conversation responses are generated from the resulting SQL tables using  Mistral-7B-Instruct-v0.2., an open-source LLM. The response generator LLM is prompted by a system prompt with a set of rules, the latest dialogue state with user intents and entities, and one or more latest user utterances along with the results from the database query. The GPT-4o model continuously assess the conversation to see if all the user goals are met and the conversation is marked as successful if the goals are met within 20 conversation turns.

\section{Results and Discussion}

In this paper, we primarily focus on the generated users and their attribute values. These values are generated by the LLM, and therefore can have inherent biases, and can be non-representative. Therefore, it is important to analyze the generated values. Discussing all attributes are beyond the scope of this paper, therefore we only choose the most important attributes, for example, demographic attributes (like gender, origin etc.), generated interests, personality attributes (personality types) etc.

\begin{figure}[!t]
  \resizebox{\columnwidth}{!}{%
    \includegraphics[width=0.5\textwidth]{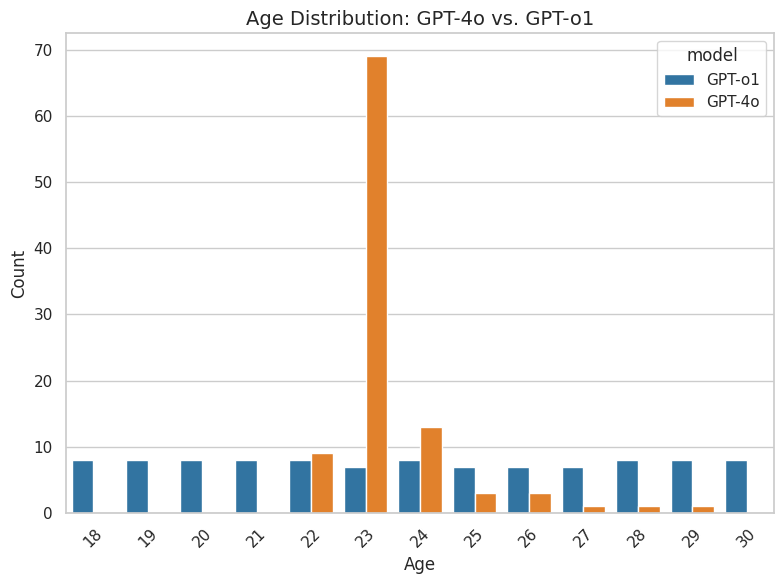} % Reduced width to 80% of the text width
  }
  \caption{User distribution of different Ages selected per model.}
  \label{fig:user_demograph_age}
\end{figure}

\subsection{Age and Desired Degree Programs}

First, we look into the age distribution of the generated users. In Figure \ref{fig:user_demograph_age}, we show the different ages chosen by each model. GPT-4o chose more users aged 22,23 and 24, while the GPT-o1 model created an equally diversified number of users across different ages. Given the fact that GPT-4o only generated users with a prospect of Masters admission as per the Table \ref{tab:desired_degree_type}, the age group seems fitting, while the  GPT-o1 enforced the equal distribution of the different user groups (which may not represent the real age distribution among students at TUB or universites in general). However, it is a potential limitation of the GPT-4o model that even if the prompt instructed to generate users with diverse degree program prospects with given four options, the model failed to do so.

\begin{table}[t]
\centering
\small
\resizebox{\columnwidth}{!}{%
\begin{tabular}{l|l|c}
\hline
\textbf{Model} & \textbf{Desired Degree} & \textbf{Count} \\
\hline

% -- MODEL 1 --
\multirow{4}{*}{\textbf{GPT-4o}}
 & Bachelor & 0 \\
 & Master & 100 \\
 & Exchange student & 0 \\
 & not sure & 0 \\
\hline

% -- MODEL 2 --
\multirow{4}{*}{\textbf{GPT-o1}}
 & Bachelor & 33 \\
 & Master & 44 \\
 & Exchange student & 14 \\
 & not sure & 9 \\
\hline

\end{tabular}
}
\caption{Desired degree types chosen by GPT-4o and GPT-o1 during user generation.}
\label{tab:desired_degree_type}
\end{table}

\begin{figure}[!t]
  \includegraphics[width=0.5\textwidth]{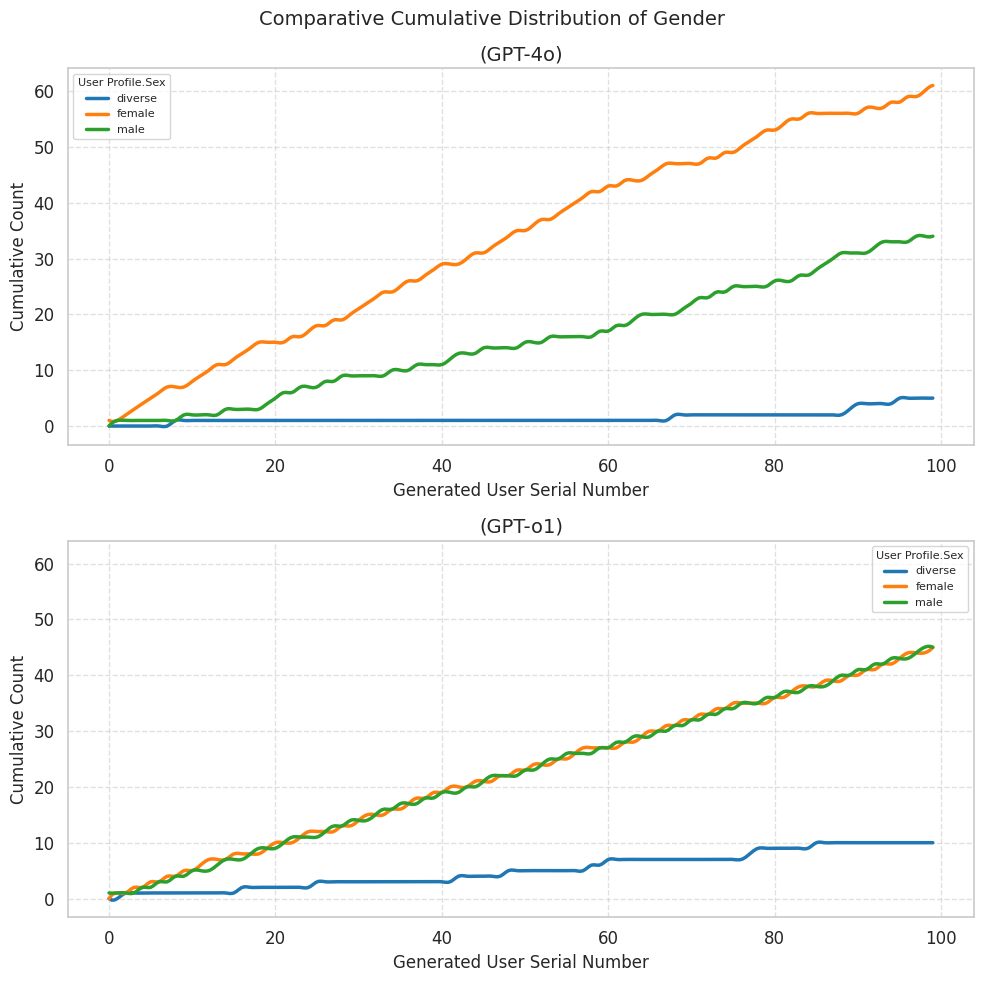} % Reduced width to 80% of the text width
  \caption{Gender selection during user simulation over time.}
  \label{fig:user_demograph_gender}
\end{figure}

\begin{table}[t]
\centering
\small
\resizebox{\columnwidth}{!}{%
\begin{tabular}{c|l|l}
\hline
\textbf{Model} & \textbf{Gender} & \textbf{Topics (Count)} \\
\hline

% ================== MODEL 1: 3 genders × 5 rows each = 15 total rows ==================
\multirow{15}{*}{\textbf{GPT-4o}}
 & \multirow{5}{*}{female}
 & urban\_planning (30) \\
 & 
 & climate\_change (24) \\
 & 
 & artificial\_intelligence (20) \\
 & 
 & sustainable\_architecture (18) \\
 & 
 & renewable\_energy (16) \\
% (No extra horizontal line here, so it appears seamless)
\cline{2-3}
 & \multirow{5}{*}{male}
 & renewable\_energy (18) \\
 & 
 & urban\_planning (11) \\
 & 
 & robotics (11) \\
 & 
 & artificial\_intelligence (11) \\
 & 
 & sustainable\_architecture (10) \\
\cline{2-3}
 & \multirow{5}{*}{diverse}
 & climate\_change (3) \\
 & 
 & sustainable\_architecture (2) \\
 & 
 & robotics (2) \\
 & 
 & artificial\_intelligence (2) \\
 & 
 & renewable\_energy (1) \\
\hline

% ================== MODEL 2: 3 genders × 5 rows each = 15 total rows ==================
\multirow{15}{*}{\textbf{GPT-o1}}
 & \multirow{5}{*}{female}
 & urban\_planning (9) \\
 & 
 & cybersecurity (8) \\
 & 
 & architecture (8) \\
 & 
 & computer\_science (7) \\
 & 
 & environmental\_engineering (6) \\
\cline{2-3}
 & \multirow{5}{*}{male}
 & telecommunications\_engineering (7) \\
 & 
 & robotics (7) \\
 & 
 & renewable\_energy (6) \\
 & 
 & industrial\_engineering (6) \\
 & 
 & mechanical\_engineering (6) \\
\cline{2-3}
 & \multirow{5}{*}{diverse}
 & architecture (3) \\
 & 
 & urban\_planning (3) \\
 & 
 & media\_informatics (3) \\
 & 
 & civil\_engineering (2) \\
 & 
 & cybersecurity (1) \\
\hline
\end{tabular}
}
\caption{Top five User Interests by Gender, GPT-4o vs GPT-o1.}
\label{tab:top5_interests_by_sex}
\end{table}

%\end{tabular}
%\caption{\small Top five user interests by gender generated by each model.}
%\label{tab:top5_interests_by_sex}
%\end{table}

\begin{figure}[!t]
  \resizebox{\columnwidth}{!}{%
  \includegraphics[width=\linewidth]{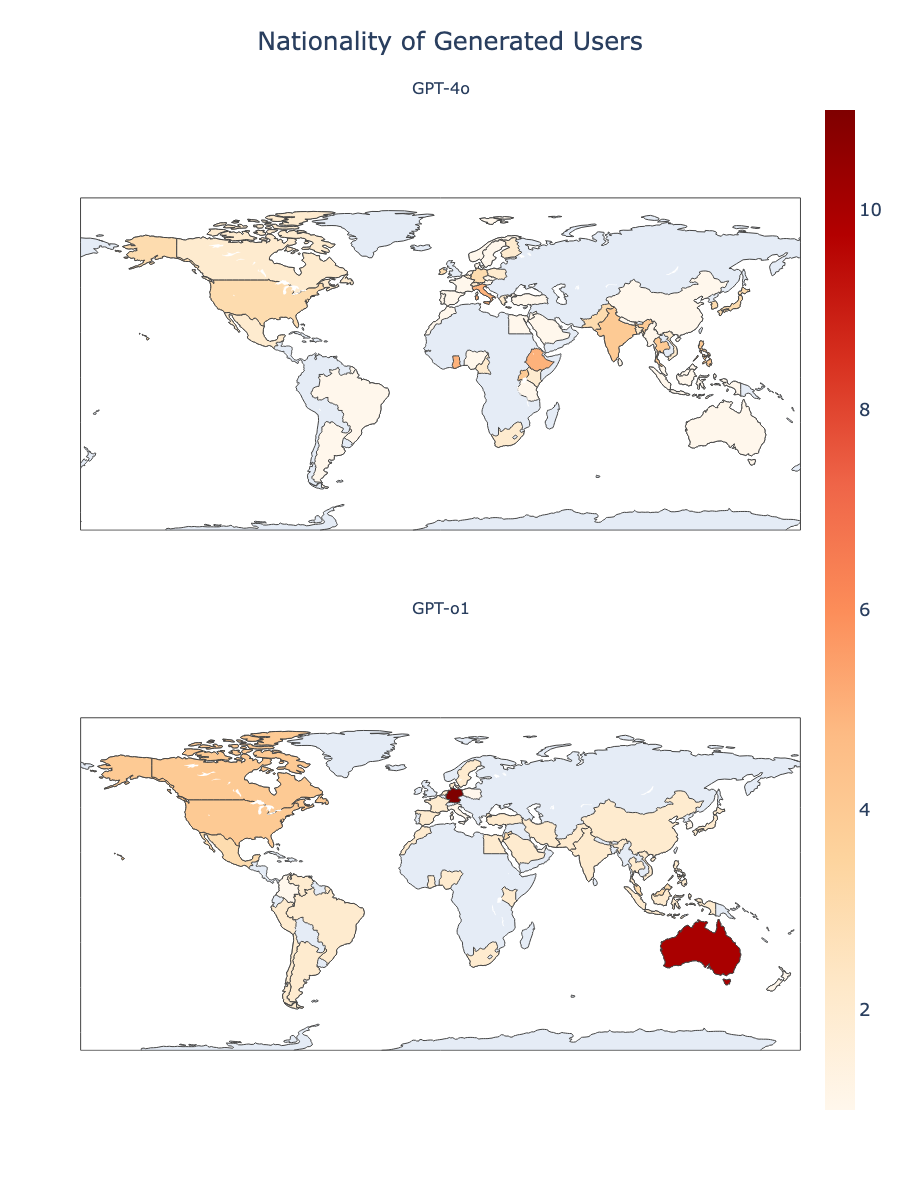} % Reduced width to 80% of the text width
  }
  \caption{User distribution from different Nationality.}
  \label{fig:user_demograph_nation}
\end{figure}

\subsection{Gender Selection}

In Figure \ref{fig:user_demograph_gender}, we plot the choice of gender by each of the models over the time during user profile generation. GPT-4o generated 61 females, 31 males and 5 users with diverse gender. On the other hand, GPT-o1 generated an equal number of males and females (45 each) and 10 diverse users. We would like to point out that, none of them represents the real demographics of the TU-Berlin studen's gender distribution, where there are 65.2\% male, 34.7\% female and 0.1\% diverse  students enrolled currently\footnote{\url{https://www.tu.berlin/en/about/profile/tu-berlin-statistics}}. When it comes to generating interests based on gender in Table \ref{tab:top5_interests_by_sex}, GPT-o1 assigned more technical subject interests for male users,  although both models generated a diverse set of interests per gender.

\begin{figure}[!t]
  \includegraphics[width=0.5\textwidth]{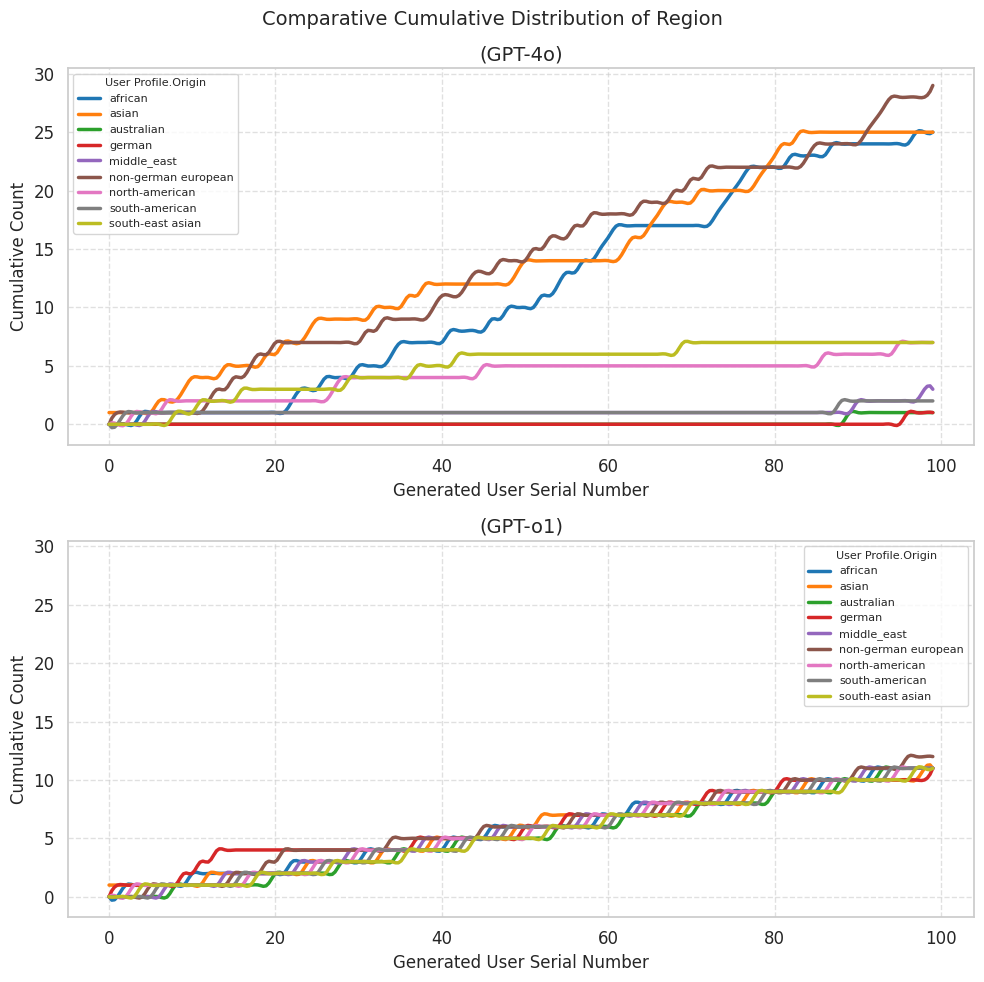} % Reduced width to 80% of the text width
  \caption{User distribution from different Regions over time.}
  \label{fig:user_demograph_region}
\end{figure}

\subsection{Region and Nationality}

In Figure \ref{fig:user_demograph_nation}, we plot a hit map based on the nationality of the generated users. GPT-4o generated more distributed user nationalities, while GPT-o1 generated more Western users. In Figure \ref{fig:user_demograph_region}, while analyzing the generated users from different regions over time, we see that GPT-4o generated more non-german European, Asian, and African users, while GPT-o1 kept a strict balance of user regions over time. We could not verify if any of these two distributions matches the original number of students in TU-berlin per region.

\begin{table}[t]
\centering
\small
\begin{tabularx}{\linewidth}{l|p{2cm}|X}
\hline
\textbf{Model} & \textbf{Region} & \textbf{Topic (Count)} \\
\hline

% ================== MODEL 1 (18 rows total) ==================
\multirow{18}{*}{\textbf{GPT-4o}}
 & \multirow{2}{*}{asian}
 & artificial\_intelligence (16) \\
 & 
 & robotics (11) \\
\cline{2-3}
 & \multirow{2}{2cm}{non-german european}
 & urban\_planning (20) \\
 & 
 & sustainable\_architecture (14) \\
\cline{2-3}
 & \multirow{2}{2cm}{south-american}
 & renewable\_energy (2) \\
 & 
 & sustainable\_architecture (2) \\
\cline{2-3}
 & \multirow{2}{*}{north-american}
 & artificial\_intelligence (3) \\
 & 
 & robotics (3) \\
\cline{2-3}
 & \multirow{2}{*}{african}
 & renewable\_energy (15) \\
 & 
 & climate\_change (14) \\
\cline{2-3}
 & \multirow{2}{*}{middle\_east}
 & climate\_change (2) \\
 & 
 & urban\_planning (1) \\
\cline{2-3}
 & \multirow{2}{*}{south-east asian}
 & artificial\_intelligence (4) \\
 & 
 & robotics (4) \\
\cline{2-3}
 & \multirow{2}{*}{australian}
 & climate\_change (1) \\
 & 
 & sustainable\_architecture (1) \\
\cline{2-3}
 & \multirow{2}{*}{german}
 & robotics (1) \\
 & 
 & artificial\_intelligence (1) \\
\hline

% ================== MODEL 2 (also 18 rows, same region order) ==================
\multirow{18}{*}{\textbf{GPT-o1}}
 & \multirow{2}{*}{asian}
 & artificial\_intelligence (6) \\
 & 
 & cybersecurity (3) \\
\cline{2-3}
 & \multirow{2}{2cm}{non-german european}
 & urban\_planning (5) \\
 & 
 & architecture (4) \\
\cline{2-3}
 & \multirow{2}{*}{south-american}
 & civil\_engineering (3) \\
 & 
 & environmental\_engineering (3) \\
\cline{2-3}
 & \multirow{2}{*}{north-american}
 & data\_science (3) \\
 & 
 & media\_informatics (3) \\
\cline{2-3}
 & \multirow{2}{*}{african}
 & industrial\_engineering (4) \\
 & 
 & architecture (3) \\
\cline{2-3}
 & \multirow{2}{*}{middle\_east}
 & mechanical\_engineering (3) \\
 & 
 & machine\_learning (2) \\
\cline{2-3}
 & \multirow{2}{*}{south-east asian}
 & urban\_planning (4) \\
 & 
 & environmental\_engineering (3) \\
\cline{2-3}
 & \multirow{2}{*}{australian}
 & media\_informatics (3) \\
 & 
 & robotics (3) \\
\cline{2-3}
 & \multirow{2}{*}{german}
 & cybersecurity (3) \\
 & 
 & robotics (2) \\
\hline

\end{tabularx}
\caption{Top two interests chosen by user's Regions, GPT-4o vs GPT-o1.}
\label{tab:top2_interests_merged}
\end{table}

In Table \ref{tab:top2_interests_merged}, we show the top two interests per region for generated users by each model. We discover that interests generated by GPT-4o match the stereotypical idea that Asian, South-east Asian, South American, and German users are more interested in information technology-related subjects (ex. Artificial intelligence and Robotics) compared to non-german European, South American, Middle-east, and Australian users who are more interested in subjects like sustainability and climate. GPT-4o is relatively diverse in assigning interest. 

\subsection{Personality Types}

\begin{figure}[!t]
  \includegraphics[width=0.5\textwidth]{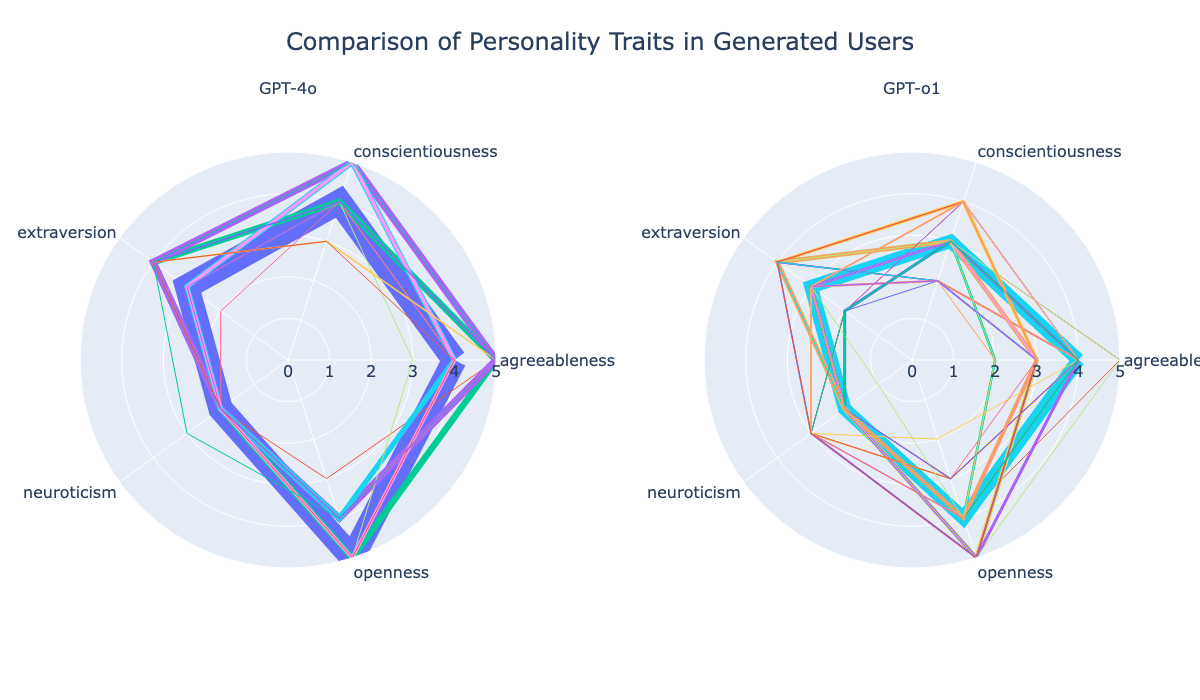} % Reduced width to 80% of the text width
  \caption{User distribution of different Personality Types per model.}
  \label{fig:user_demograph_personality}
\end{figure}

For user interaction generation, it is important to define the personality type and generate users with diverse personality types. We use five-factor \cite{Chmielewski2013} personality model, which consists of five personality factors across 5 different dimensions. We ask the LLMs to assign a value between 1-5 for each of these personality dimensions. Figure \ref{fig:user_demograph_personality} shows that both models create users with less Neuroticism. Neuroticism is one of the Big Five personality traits that affect people most negatively if they have too much of it. The GPT-4o model assigns more extreme values in Conscientiousness, Agreeableness, Extraversion and Openness dimensions compared to the GPT-o1 model, which sets moderate values across these dimensions.

From the generated users, it is evident that the GPT-4o model creates users with more variations, while the GPT-o1 model enforces balance in distributions (ex. user region, user gender, age, etc.). Both models create a diverse set of user interests, but GPT-4o shows some stereotypicality when choosing interests for users based on region (which may reflect the real-world distribution). Both models create a diverse set of personality types, with GPT-4o creating users with more extreme values in personality dimensions. A more interesting analysis would be to see the personality types based on the region, but that is beyond the scope of the paper.

\subsection{Conversation Simulation}

We used 57 users generated using GPT-o1 to simulate conversations with StudyBot. Each conversation has a maximum duration of 20 turns and the conversation is terminated earlier if all the set user goals are achieved. We use GPT-4o to generate the user utterances given a prompt and the user template. StudyBot recognizes the user's intent, creates and executes SQL queries in the database, and generates a natural language response using Mistral-7b.v2 from the SQL-query result.

Among the conversations, 82.46\% sessions achieved all the user goals as judged by the GPT-4o LLMs. The average length of per-user utterance for users with formal communication type is 12.11 words, while for the informal communication type, it is 12.71 words. Although this doesn't show a significant difference, increasing the number of conversations might lead to a significant difference. The average number of conversation turns for the sessions where the conversation goals are achieved is 10.94. Table \ref{tab:secondary_goals_combined_tb} shows the top three goals for which the conversations were successful and unsuccessful. This gives us a better idea about what questions StudyBot is struggling to answer. 

\begin{table}[h]
    \centering
    \begin{tabular}{l c}
        \hline
        \multicolumn{2}{c}{\textbf{ goal\_achieved=True}} \\
        \hline
        \textbf{Secondary Goal} & \textbf{Count} \\
        \hline
        module\_contents        & 20 \\
        acquired\_skills        & 16 \\
        admission\_requirements & 15 \\
        \hline
        \\[-1em] % Adjust spacing between sections if needed
        \hline
        \multicolumn{2}{c}{\textbf{goal\_achieved=False}} \\
        \hline
        \textbf{Secondary Goal} & \textbf{Count} \\
        \hline
        admission\_restriction     & 4 \\
        module\_contents           & 3 \\
        structure\_of\_the\_program & 3 \\
        \hline
    \end{tabular}
\caption{Top three Secondary Goals with goal\_achieved=True and goal\_achieved=False.}
\label{tab:secondary_goals_combined_tb}
\end{table}

The same goes for table \ref{tab:general_interests_combined_tb}, which shows for what user interests the chatbot is being successful or failing to provide appropriate responses. All these evaluations are being done automatically without any human supervision.

\begin{table}[h]
    \centering
    \begin{tabular}{l c}
        \hline
        \multicolumn{2}{c}{\textbf{goal\_achieved=True}} \\
        \hline
        \textbf{General Interest} & \textbf{Count} \\
        \hline
        renewable\_energy               & 7 \\
        environmental\_engineering      & 7 \\
        telecommunications\_engineering & 6 \\
        \hline
        \\[-1em] % Adjust spacing between sections if needed
        \hline
        \multicolumn{2}{c}{\textbf{goal\_achieved=False}} \\
        \hline
        \textbf{General Interest} & \textbf{Count} \\
        \hline
        software\_engineering   & 3 \\
        computer\_science       & 3 \\
        electrical\_engineering & 2 \\
        \hline
    \end{tabular}
    \caption{Top three General Interests with goal\_achieved=True and goal\_achieved=False.}
    \label{tab:general_interests_combined_tb}
\end{table}

\section{Conclusion}

In this study, we introduced a novel user simulation approach utilizing Large Language Models (LLMs) to generate diverse user personas and engage them in simulated dialogues with a task-oriented chatbot, StudyBot. Our approach demonstrated the feasibility of automating user profile generation, conversation simulation, and evaluation processes with minimal human intervention. We leveraged GPT-4o and GPT-o1 to generate synthetic users with varied demographics, interests, conversational styles, and personality traits, which were then used to assess the performance of StudyBot.

Our analysis revealed several key insights. First, GPT-4o and GPT-o1 differed in their approach to user generation, with GPT-4o producing users with more varied attributes, while GPT-o1 enforced a more balanced distribution. In the dialogue simulations using generated users, we observed that StudyBot successfully met user-defined goals in 82.46\% of the dialogue sessions.

The results highlight the potential of a LLM-driven user simulation as an effective tool for evaluating and improving task-oriented dialogue systems. By automating user generation and conversation evaluation, our approach enables rapid, scalable testing of dialogue systems while reducing reliance on manually curated datasets. Future work could explore refining LLM-generated user diversity, mitigating potential biases, and expanding the conversational scope to assess more complex multi-turn interactions.

\section*{Acknowledgments}

Parts of the presented work and this paper have been funded by the Federal Ministry of Education and Research (Germany) and the Federal State of Berlin under grant no. 16DHBKI088 for the project USOS at Technische Universität Berlin.

% Bibliography entries for the entire Anthology, followed by custom entries
%\bibliography{anthology,custom}
% Custom bibliography entries only
\bibliography{main}

\begin{thebibliography}{10}
\providecommand{\natexlab}[1]{#1}

\bibitem[{Abdullin et~al.(2024)Abdullin, Molla-Aliod, Ofoghi, Yearwood, and
  Li}]{abdullin2024synthetic}
Yelaman Abdullin, Diego Molla-Aliod, Bahadorreza Ofoghi, John Yearwood, and
  Qingyang Li. 2024.
\newblock Synthetic dialogue dataset generation using llm agents.
\newblock \emph{arXiv preprint arXiv:2401.17461}.

\bibitem[{Achiam et~al.(2023)Achiam, Adler, Agarwal, Ahmad, Akkaya, Aleman,
  Almeida, Altenschmidt, Altman, Anadkat et~al.}]{openai2024gpt4}
Josh Achiam, Steven Adler, Sandhini Agarwal, Lama Ahmad, Ilge Akkaya,
  Florencia~Leoni Aleman, Diogo Almeida, Janko Altenschmidt, Sam Altman,
  Shyamal Anadkat, et~al. 2023.
\newblock Gpt-4 technical report.
\newblock \emph{arXiv preprint arXiv:2303.08774}.

\bibitem[{Brown et~al.(2020)Brown, Mann, Ryder, Subbiah, Kaplan, Dhariwal,
  Neelakantan, Shyam, Sastry, Askell et~al.}]{brown2020language}
Tom Brown, Benjamin Mann, Nick Ryder, Melanie Subbiah, Jared~D Kaplan, Prafulla
  Dhariwal, Arvind Neelakantan, Pranav Shyam, Girish Sastry, Amanda Askell,
  et~al. 2020.
\newblock Language models are few-shot learners.
\newblock \emph{Advances in neural information processing systems},
  33:1877--1901.

\bibitem[{Chmielewski and Morgan(2013)}]{Chmielewski2013}
Michael~S. Chmielewski and Theresa~A. Morgan. 2013.
\newblock \href {https://doi.org/10.1007/978-1-4419-1005-9_1226}
  {\emph{Five-Factor Model of Personality}}, pages 803--804.
\newblock Springer New York, New York, NY.

\bibitem[{G{\"u}r et~al.(2018)G{\"u}r, Hakkani-T{\"u}r, T{\"u}r, and
  Shah}]{gur2018user}
Izzeddin G{\"u}r, Dilek Hakkani-T{\"u}r, Gokhan T{\"u}r, and Pararth Shah.
  2018.
\newblock User modeling for task oriented dialogues.
\newblock In \emph{2018 IEEE Spoken Language Technology Workshop (SLT)}, pages
  900--906. IEEE.

\bibitem[{Hude{\v{c}}ek and Du{\v{s}}ek(2023)}]{hudevcek2023large}
Vojt{\v{e}}ch Hude{\v{c}}ek and Ond{\v{r}}ej Du{\v{s}}ek. 2023.
\newblock Are large language models all you need for task-oriented dialogue?
\newblock In \emph{Proceedings of the 24th Annual Meeting of the Special
  Interest Group on Discourse and Dialogue}, pages 216--228.

\bibitem[{Jiang et~al.(2023)Jiang, Sablayrolles, Mensch, Bamford, Chaplot,
  de~las Casas, Bressand, Lengyel, Lample, Saulnier, Lavaud, Lachaux, Stock,
  Scao, Lavril, Wang, Lacroix, and Sayed}]{jiang2023mistral}
Albert~Q. Jiang, Alexandre Sablayrolles, Arthur Mensch, Chris Bamford,
  Devendra~Singh Chaplot, Diego de~las Casas, Florian Bressand, Gianna Lengyel,
  Guillaume Lample, Lucile Saulnier, Lélio~Renard Lavaud, Marie-Anne Lachaux,
  Pierre Stock, Teven~Le Scao, Thibaut Lavril, Thomas Wang, Timothée Lacroix,
  and William~El Sayed. 2023.
\newblock \href {https://arxiv.org/abs/2310.06825} {Mistral 7b}.
\newblock \emph{Preprint}, arXiv:2310.06825.

\bibitem[{Schatzmann and Young(2009)}]{schatzmann2009hidden}
Jost Schatzmann and Steve Young. 2009.
\newblock The hidden agenda user simulation model.
\newblock \emph{IEEE transactions on audio, speech, and language processing},
  17(4):733--747.

\bibitem[{Wei et~al.(2021)Wei, Bosma, Zhao, Guu, Yu, Lester, Du, Dai, and
  Le}]{wei2021finetuned}
Jason Wei, Maarten Bosma, Vincent Zhao, Kelvin Guu, Adams~Wei Yu, Brian Lester,
  Nan Du, Andrew~M Dai, and Quoc~V Le. 2021.
\newblock Finetuned language models are zero-shot learners.
\newblock In \emph{International Conference on Learning Representations}.

\bibitem[{Zhao et~al.(2019)Zhao, Li, and Lin}]{zhao2019review}
Yin~Jiang Zhao, Yan~Ling Li, and Min Lin. 2019.
\newblock A review of the research on dialogue management of task-oriented
  systems.
\newblock In \emph{Journal of Physics: Conference Series}, volume 1267, page
  012025. IOP Publishing.

\end{thebibliography}
\newpage

\appendix
\section{Appendix}

\subsection{User Template}
An example of a generated user is added in Figure \ref{fig:user}. It also contains the attributes that are being used in the user generation template.

\subsection{System Prompt}

\begin{figure}[!t]
  \resizebox{\columnwidth}{!}{%
  \includegraphics[width=0.5\textwidth]{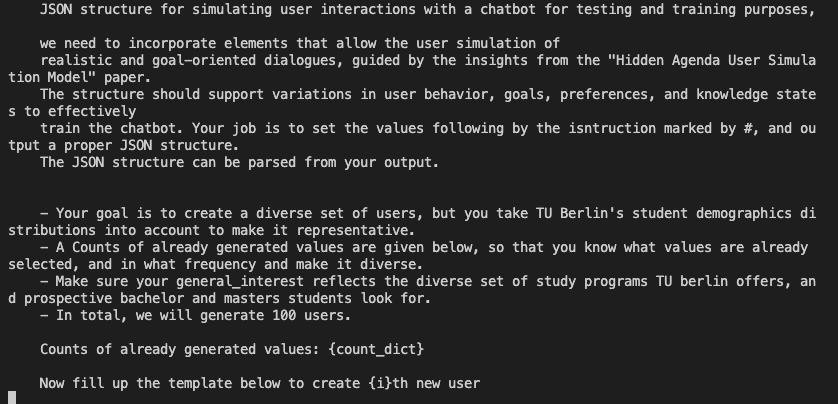} % Reduced width to 80% of the text width
  }
  \caption{System prompt for user simulation.}
  \label{fig:system}
\end{figure}

The system prompt to generate the user is given in Figure \ref{fig:system}, where the model was instructed to generate a diverse set of users given the currently generated user statistics and a user template.

\subsection{Conversation Simulation}

\begin{figure*}
  \centering
  \includegraphics[width=\textwidth, height=\textheight, keepaspectratio]{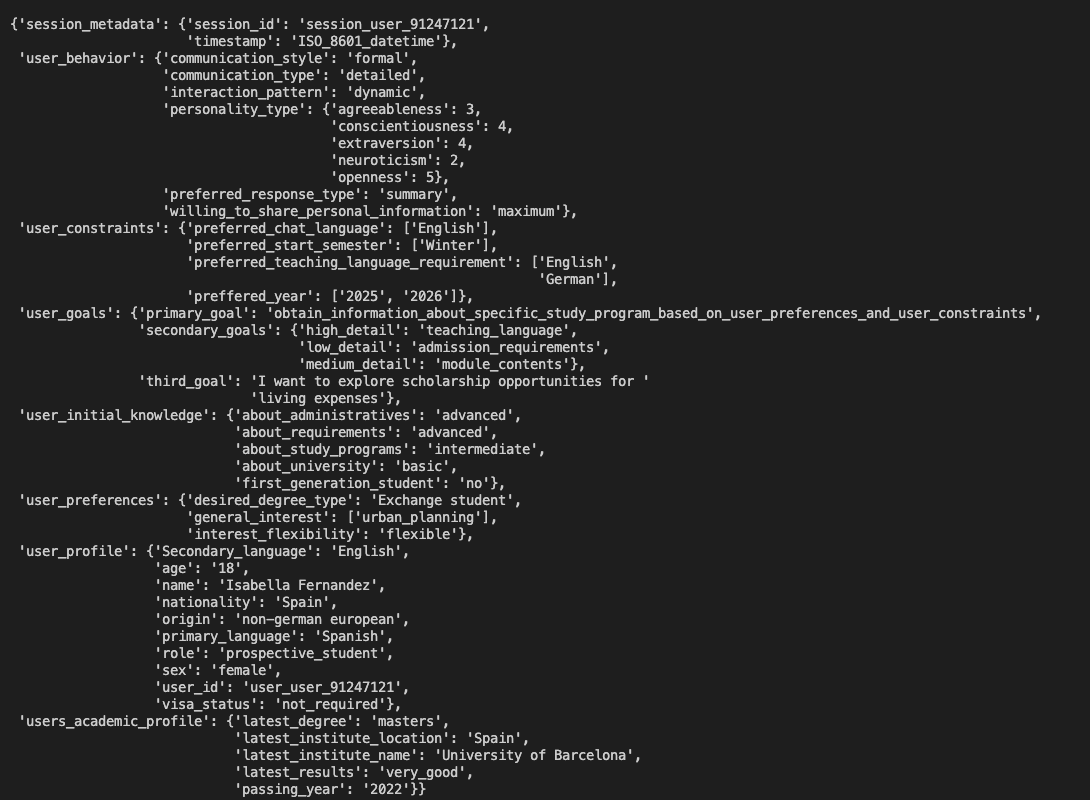} % Image scales to fill the page but maintains aspect ratio
  \caption{Example of a generated user (generated using GPT-o1).}
  \label{fig:user}
\end{figure*}

\begin{figure*}
  \centering
  \includegraphics[width=\textwidth, height=\textheight, keepaspectratio]{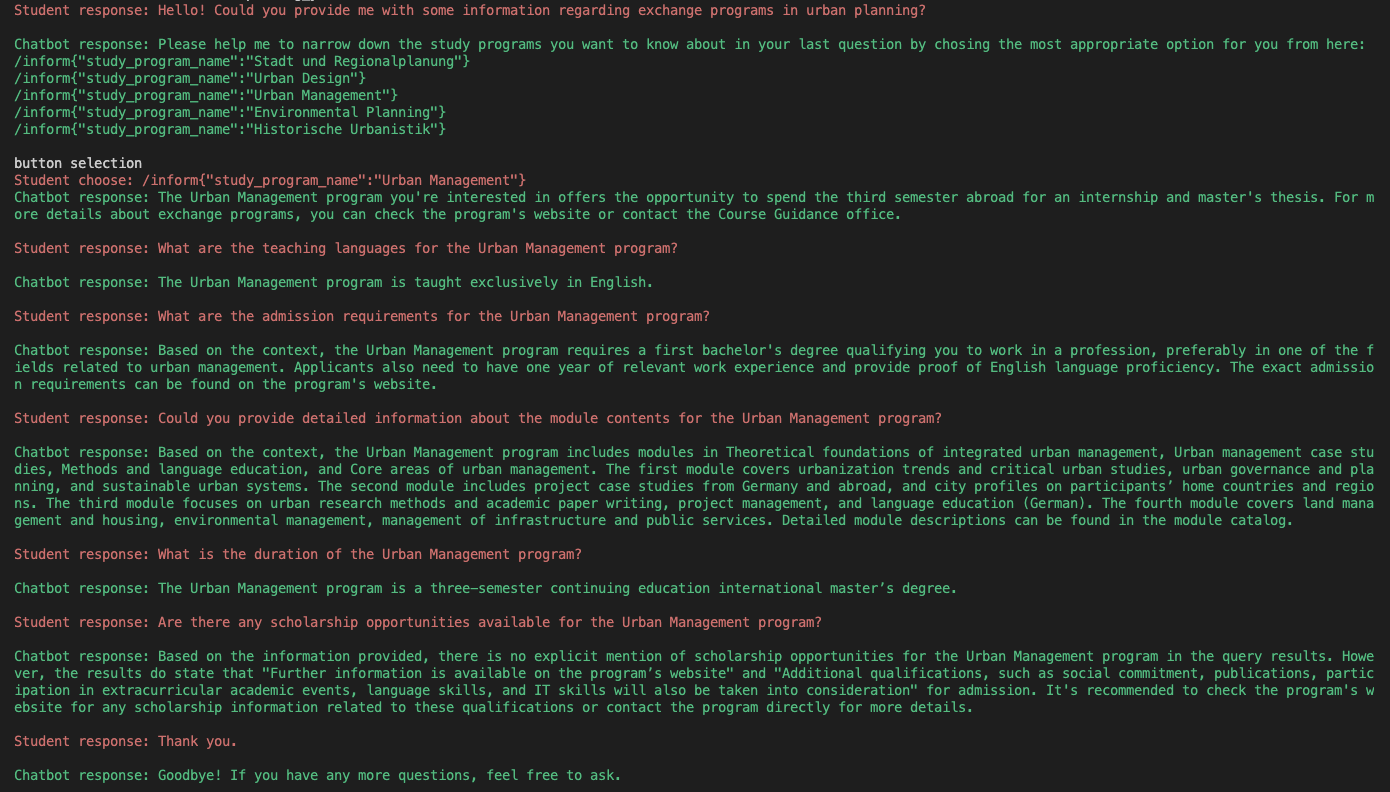} % Image scales to fill the page but maintains aspect ratio
  \caption{Example of dialogue session with StudyBot using simulated user.}
  \label{fig:chat}
\end{figure*}

An example of conversation session with StudyBot using the user in Figure \ref{fig:user} is given in Figure \ref{fig:chat}. The red is marked as user utterance and the green is marked as system utterances. This particular conversation is marked as successful as all the user goals defined in `user\_goals` are met.

\label{sec:appendix}

\end{document}